\documentclass[sigconf]{acmart}
\usepackage{libertine}
\usepackage{array}
\usepackage{multirow}

\AtBeginDocument{%
  }

\copyrightyear{2025}
\acmYear{2025}
\setcopyright{acmlicensed}
\acmConference[MM '25]{Proceedings of the 33rd ACM International Conference on Multimedia}{October 27--31, 2025}{Dublin, Ireland}
\acmBooktitle{Proceedings of the 33rd ACM International Conference on Multimedia (MM '25), October 27--31, 2025, Dublin, Ireland}
\acmDOI{10.1145/3746027.3755674}
\acmISBN{979-8-4007-2035-2/2025/10}

\begin{document}

\title{I2CR: Intra- and Inter-modal Collaborative Reflections for Multimodal Entity Linking}

% \author{Anonymous Authors}

% \affiliation{%

\author{Ziyan Liu}
\affiliation{%
  \institution{East China University of Science and Technology}
  \city{Shanghai}
  \country{China}}
\email{ y30241069@mail.ecust.edu.cn}

\author{Junwen Li}
\affiliation{%
  \institution{East China University of Science and Technology}
  \city{Shanghai}
  \country{China}}
\email{ 18602126280@163.com}

\author{Kaiwen Li}
\affiliation{%
  \institution{South China University of Technology}
  \city{Guangzhou}
  \country{China}}
\email{202164030293@mail.scut.edu.cn}

\author{Tong Ruan}
\affiliation{%
  \institution{East China University of Science and Technology}
  \city{Shanghai}
  \country{China}}
\email{ruantong@ecust.edu.cn}

\author{Chao Wang}
\affiliation{%
  \institution{Shanghai University}
  \city{Shanghai}
  \country{China}}
\email{cwang@shu.edu.cn}

\author{Xinyan He}
\affiliation{%
  \institution{Meituan}
  \city{Shanghai}
  \country{China}}
\email{hexinyan03@meituan.com}

\author{Zongyu Wang}
\affiliation{%
  \institution{Meituan}
  \city{Shanghai}
  \country{China}}
\email{ wangzongyu02@meituan.com}

\author{Xuezhi Cao}
\affiliation{%
  \institution{Meituan}
  \city{Shanghai}
  \country{China}}
\email{ caoxuezhi@meituan.com}

\author{Jingping Liu}
\authornote{Corresponding author.}
\affiliation{%
  \institution{East China University of Science and Technology}
  \city{Shanghai}
  \country{China}}
\email{ jingpingliu@ecust.edu.cn}

\renewcommand{\shortauthors}{Ziyan Liu et al.}

\begin{abstract}
Multimodal entity linking plays a crucial role in a wide range of applications. Recent advances in large language model-based methods have become the dominant paradigm for this task, effectively leveraging both textual and visual modalities to enhance performance.  Despite their success, these methods still face two challenges, including unnecessary incorporation of image data in certain scenarios and the reliance only on a one-time extraction of visual features, which can undermine their effectiveness and accuracy. To address these challenges, we propose a novel LLM-based framework for the multimodal entity linking task, called Intra- and Inter-modal Collaborative Reflections. 
This framework prioritizes leveraging text information to address the task. When text alone is insufficient to link the correct entity through intra- and inter-modality evaluations, it employs a multi-round iterative strategy that integrates key visual clues from various aspects of the image to support reasoning and enhance matching accuracy. Extensive experiments on three widely used public datasets demonstrate that our framework consistently outperforms current state-of-the-art methods in the task, achieving improvements of 3.2\%, 5.1\%, and 1.6\%, respectively. Our code is available at \url{https://github.com/ziyan-xiaoyu/I2CR/}.

\end{abstract}





\maketitle

\section{Introduction}

Entity linking\cite{shen2014entity, shen2021entity, chen2023end}, which maps ambiguous mentions in text to the standard entities in a knowledge graph (KG)\cite{chen2024knowledge}, e.g., Wikidata, is essential for applications such as question answering \cite{shah2019kvqa, longpre2021entity}, and recommendation systems \cite{deldjoo2020recommender}. As multimodal contexts (images and text) become more common in real-world scenarios, recent studies \cite{wang2022wikidiverse, yao2023ameli, liu2024beyond} suggest incorporating images to enhance entity disambiguation, leading to the emergence of Multimodal Entity Linking (MEL)\cite{gan2021multimodal}. 

% 1. 目前，主流的解决MEL任务的方法可以分为两大类。
% 2. 一是基于传统深度学习的方法，首先通过分别对mention的文本和图像提取特征并进行融合\cite{luo2023multi, song2024dual, song2024dwe+}，或使用交叉注意力机制直接提取多模态特征\cite{dongjie2022multimodal, zhang2024optimal}；然后，将融合特征与实体特征进行匹配，选择匹配度最高的实体。
% 3. 但是这类方法存在：1）模型缺乏足够的先验知识，导致其难以处理需要复杂推理的样本；2）模型泛化能力不强，导致其难以适应新场景或未见过的实体。
% 4. 二是基于大语言模型（LLM）生成的方法。
% 5. 这类方法主要是将通过小模型提取的视觉特征（或者图像）和文本同时输入给LLM（或者multi-modal LLM （MLLM）），并让模型从KG中为给定的mention选择合适实体\cite{shi2024generative,liu2024unimel}。
% 6. 尽管大模型能凭借其强大的知识储备、知识推理以及文本生成能力有效地弥补传统深度学习方法先验知识不足和泛化能力不强的问题，但仍存在以下两个问题：
% 7. 1）在实体链接任务中，并非所有场景都需要额外的图像信息，例如，当文本描述已经足够清晰时，引入图像可能不仅无法提供有效的辅助信息，反而可能带来噪声，从而降低匹配精度。此外，额外的视觉处理也会增加计算成本，降低推理效率。
% 8. 2）现有方法通常只用一个模块来从图像提取视觉特征或获取其文本描述。但是，一个模块获取的信息往往是不够充分的，导致模型难以准确的为mention选择合适的实体。

\begin{figure}[t]
\centering
\includegraphics[width=0.9\linewidth]{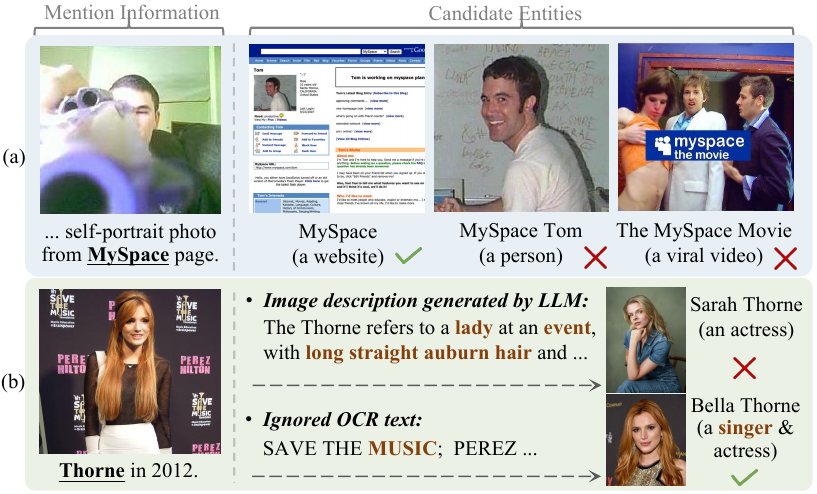}
\caption{Two limitations of current LLM-based methods for the MEL task: (a) unnecessary incorporation of image data, and (b) reliance on a one-time extraction of visual features.}
\label{fig:introduction}
\end{figure}
% in certain cases
In the early stages, numerous deep learning (DL)-based\cite{lecun2015deep} methods have been developed for the MEL task. These methods generally extract unimodal features for the text and image separately, followed by feature fusion \cite{luo2023multi, song2024dual, song2024dwe+}, or directly extract multimodal features from the text and image through cross-attention mechanisms \cite{dongjie2022multimodal, zhang2024optimal}. These extracted features are then compared with entity features to identify the target that best matches the given mention. However, such methods face two challenges: 1) The models lack sufficient prior knowledge, making it difficult to handle samples requiring deeper reasoning. 
For instance, given the mention ``Presidential State Car'', DL-based methods might identify the car in the mention image as a \textit{Cadillac limousine}. However, without knowing that this type of car is used by the U.S. president, it may not link it to the correct entity ``Presidential state car (United States)''.
2) These models also don’t generalize well because they’re trained on one specific task with limited data. They often learn patterns that only work for the training examples. So, if the input changes even slightly, the model might get confused. Instead of learning how to deal with new situations, it mostly just memorizes the training data.

% The model’s poor generalization ability hampers its adaptability to new scenarios or unseen entities. 

% 1. 比如说，在数据集WikiMEL上训练的传统深度学习模型直接在另一个数据集WikiDiverse（its entity types远比WikiMEL的entity types丰富的多）上测试，效果会有急剧的下降。

With the rise of large language models (LLMs), LLM-based methods have become the mainstream paradigm for solving MEL tasks. These methods typically feed both text and images (or its visual features extracted by an image encoder) into the multimodal LLM (MLLM), enabling the model to select the most relevant entity from the KG for a given mention\cite{shi2024generative,liu2024unimel}. 
Since LLMs are trained on vast corpora, they possess extensive knowledge and robust generalization capabilities, which allow them to overcome the issues of insufficient prior knowledge and limited generalization faced by traditional DL-based MEL methods. However, two main challenges remain: 1) In the MEL task, not all scenarios require additional image information. For instance, in Figure \ref{fig:introduction}(a), the model can correctly infer that the mention ``MySpace'' refers to a website based on the textual context alone. However, when the image is introduced (showing a person holding a gun), it may mislead the model into selecting ``a person'' or ``a viral video'' instead of the correct entity.
2) Existing methods typically rely on a single-pass approach to extract visual features from an image or generate its text description. However, this one-time processing often fails to capture the full range of relevant information, leading to incomplete or inaccurate understanding of the given mention image.
As shown in Figure \ref{fig:introduction}(b), if the model relies solely on the description generated by the LLM, it may incorrectly identify the mention ``Thorne'' as an actress. However, if the OCR result from the image, which shows the word ``MUSIC'', is also considered, the model can more accurately infer that ``Thorne'' is not only an actress but also a singer. 
% \textcolor{red}{This highlights the importance of integrating complementary visual clues to improve semantic understanding in the task.}
% , as illustrated in Figure \ref{fig:introduction}, hindering the model’s ability to accurately select the appropriate entity for the mention.

% 1. To address the above problems, this paper introduces an innovative approach known as LLM-based Intra- and Inter-modal  Collaboration Reflections (denoted as I2CR) for MEL task.
% 2. I2CR优先利用文本信息来解决问题，对于仅依靠文本上下文无法解决的问题，再进一步采用多轮迭代的策略引入图像中不同方面的关键线索，以辅助推理并提升匹配的准确性。
% 3. Initially，我们使用字面相关性匹配方法和经过MEL数据微调的LLM to select a candidate entity from the KG for a given mention. 
% 4. Then, intra-modal consistency reflection is designed to evaluate the semantic granularity between the entity and mention, thereby determining the necessity of re-selection.   
% 5. Finally, I2CR introduces a modality alignment reflection, involving inter-modal alignment verification and visual iterative feedback, to decide if further selection based on visual clues is required.
% 6. In scenarios necessitating image clues, we employ several rounds of iteration invoking various image-to-text models, fully exploiting images information from diverse perspectives and avoiding information overload.

% multi-facted
To address the above problems, we propose an innovative framework known as LLM-based Intra- and Inter-modal Collaborative Reflections (denoted as I2CR) for the MEL task. 
It primarily relies on text information to solve problems, then verifies the results using intra- and inter-modal strategies, and finally introduces visual clues through a multi-round iterative process when text alone is insufficient.
% \textcolor{red}{I2CR primarily relies on text information to solve problems and uses a multi-round iterative strategy to introduce visual clues when text alone is not enough, helping to improve reasoning and matching accuracy.} 
Specifically, we first utilize the lexical relevance matching method and a LLM fine-tuned on MEL data to select a candidate entity from the KG for the given mention. To ensure the selection is faithful to the mention context, we then design intra-modal consistency reflection, which evaluates the semantic granularity between the mention and the selected entity. If the alignment is weak, this step triggers entity re-selection.
Finally, we propose an innovative utilization of images through a modality alignment reflection strategy. This includes inter-modal alignment check and visual iterative feedback to decide whether visual information is necessary for refining the selection. When image input is required, we perform multiple rounds of iteration using various image-to-text models, leveraging visual clues from different perspectives while avoiding information overload. Unlike methods that directly fuse text and image modalities, our strategy simplifies the task and reduces noise introduced to the LLM.

\textbf{Contributions.} The contributions of this study are summarized as follows:
\begin{itemize}
    \item We propose a novel I2CR framework that primarily relies on text information and integrates visual clues through a multi-round iterative process only when text alone is insufficient, reducing images noise introduced to the LLM.

    \item We propose a novel approach for image utilization that leverages the visual modality to verify text-based results and transforms image information into diverse textual visual cues, improving the entity matching performance of LLMs.

    % Using visual modality to verify textual results and iteratively integrating image clues when text clues are partially absent. \textcolor{red}{This approach simplifies the complexity of fusing image and text information.}

    \item Experimental results demonstrate that our model achieves state-of-the-art (SoTA) performance on three widely used datasets, with a top-1 accuracy of 92.2\% (+3.2\%) on WikiMEL, 91.6\% (+5.1\%) on WikiDiverse, and 86.8\% (+1.6\%) on RichMEL. 
\end{itemize}

\section{Related Work}

Related work in this study can be divided into three categories: entity linking, multimodal entity linking, and LLM-based reflection.

\textbf{Entity linking.}
Most previous methods for entity linking using only text follow two stages: retrieval\cite{de2020autoregressive} followed by re-ranking\cite{sil2013re, barba2022extend}. 
In the retrieval stage, the system searches a large-scale KG to find entities related to the mention. In the re-ranking stage, it scores and ranks these candidates to choose the best one.
For instance, BLINK \cite{wu2020scalable} uses a bi-encoder\cite{zhang2022optimizing} to represent the context of the mention and the descriptions of entities to retrieve candidates. It then uses a cross-encoder to rank them.
Similarly, Xu et al. \cite{xu2022enhancing} use dual encoders for retrieval and a LUKE-based cross-encoder for ranking. EPGEL \cite{lai2022improving} builds an entity profile and uses a seq2seq model\cite{sriram2017cold, egonmwan2019transformer} to rank candidates.
However, when the text is brief or vague, it’s difficult to link entities accurately using only text\cite{adjali2020multimodal}. In these cases, adding image information can help improve the results.

\textbf{Multimodal entity linking.}
% Existing approaches in the MEL task can be broadly categorized into two groups: traditional DL- and LLM-based methods. 
Existing approaches for MEL generally fall into two categories: traditional DL- and LLM-based methods.
% 1. 在第一类中，一些学者提出了分别对mention text and image进行编码，并设计了一系列融合方法将获得的特征进行合并，并与候选实体进行匹配。
% In the first category, many researchers propose encoding the mention text and image separately, then designing fusion methods to combine the features and match them with candidate entities. For example, MIMIC \cite{luo2023multi} uses textual and visual encoders to extract both global and local features, then combines them using three interaction units and contrastive learning to ensure consistency. DWE \cite{song2024dual} and DWE+ \cite{song2024dwe+} focus on refining visual features using fine-grained image attributes and developing cross-modal enhancers to bridge the semantic gap between mention text and image information.
% Other researchers extract multimodal features directly from text and images. For instance, GHMFC \cite{wang2022multimodal} proposes a co-attention mechanism to capture fine-grained inter-modal context, DZMNED \cite{moon2018multimodal} introduces modality attention for multi-modal feature extraction, and OT-MEL \cite{zhang2024optimal} formulates correlation assignment as an optimal transport problem to improve multimodal fusion.
% many approaches involve encoding the mention text and image separately, followed by fusion methods to match features with candidate entities. 
% ----------revised:
% In the first group, traditional DL-based approaches often involve encoding the mention text and image separately, followed by various fusion techniques to align these features with candidate entities. 
In the first group, traditional DL-based methods usually encode the text and image of a mention separately, then combine these features to align them with candidate entities.
% For instance, MIMIC \cite{luo2023multi} extracts global descriptive feature and local detailed features using separate text and image encoders for mentions and entities. It then uses three interaction units to calculate a matching score for each mention-entity pair.
For instance, MIMIC \cite{luo2023multi} extracts both global and local features from text and images using separate encoders, then uses three interaction units to calculate a matching score for each mention-entity pair.
% Similarly, DWE \cite{song2024dual} and DWE+ \cite{song2024dwe+} enhance visual features by incorporating fine-grained image attributes (e.g., scene feature) and applying cross-modal enhancers to improve alignment between the textual and visual modalities.
Similarly, DWE \cite{song2024dual} and DWE+ \cite{song2024dwe+} enhance visual features by adding fine-grained image details (e.g., scene feature) and applying cross-modal enhancers to better align text and image modalities.
Other approaches directly extract multimodal features from text and images, such as DZMNED employing modality attention \cite{moon2018multimodal}, GHMFC using gated fusion and contrastive training \cite{wang2022multimodal}, and OT-MEL modeling fusion as an optimal transport problem \cite{zhang2024optimal}. 
% In the second group, GEMEL \cite{shi2024generative} introduces a generative multimodal entity linking method that directly generates the target entity name for a given mention by training a feature mapper to enable cross-modal interactions.
In the second group, GEMEL \cite{shi2024generative} introduces a generative MEL method that directly generates the target entity name for a given mention by training a feature mapper to enable cross-modal interactions.
% UniMEL \cite{liu2024unimel} leverages LLMs to enhance the representation of mentions and entities by incorporating both textual and visual information. It then uses LLMs to select the target entity from the candidates with only a small number of parameters fine-tuned.
UniMEL \cite{liu2024unimel} leverages LLMs to enhance the representation of mentions and entities by incorporating both textual and visual information. It then uses LLMs to select the target entity from the candidates with only minimal fine-tuned.
In this study, we adopt the second paradigm. However, as mentioned earlier, current LLM-based methods unnecessarily incorporate image data in some cases and rely only on a one-time extraction of visual features, which may limit their effectiveness.
% 在本文，我们follow第二类的方法。但是正如前面所示，目前LLM-based method仍存在两个问题：并非所有场景都需要图像信息，过度引入可能带来噪声并增加计算成本，降低推理效率。此外，现有方法常将视觉信息简化为单一特征或描述，未能充分捕捉图像的多维信息，影响推理准确性。

\begin{figure*}[t]
\centering
\includegraphics[width=0.94\linewidth]{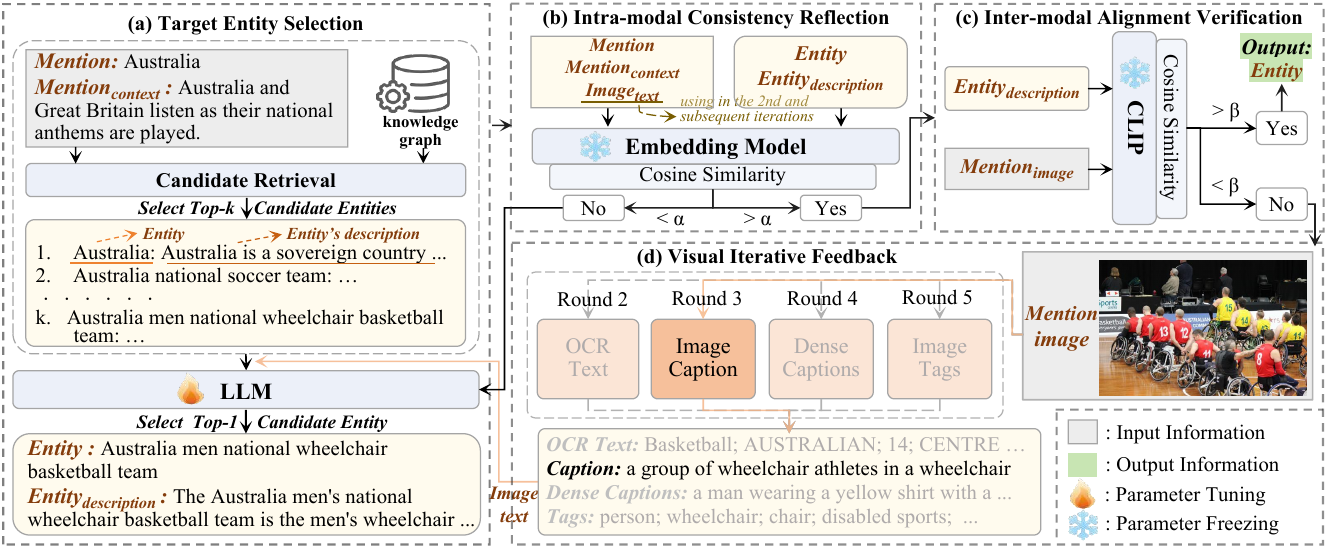}
\caption{Our I2CR framework for the MEL task.}
\label{fig:framework}
\end{figure*}

\textbf{LLM-based reflection.}
% LLMs have been widely adopted for a broad range of tasks due to their impressive capabilities. However, their performance is often limited by challenges such as hallucinations, where they generate inaccurate or fabricated information, and unfaithful reasoning. To address these issues, one promising solution is the incorporation of reflection steps \cite{pan2023automatically}, which allow the models to self-refine their outputs. 
LLMs have been widely adopted for various tasks due to their impressive capabilities. However, they still face challenges like hallucinations—producing inaccurate or fabricated content—and unfaithful reasoning. A promising solution is to incorporate reflection steps \cite{pan2023automatically}, allowing models to self-refine their own outputs.
% The sources of feedback for the reflection can be broadly categorized into two types: 1) Self-provided feedback generated by the LLM itself \cite{shinn2023reflexion} and 2) Feedback injected through external means \cite{peng2023check}.
Reflection feedback can be divided into two main types: 1) Self-provided feedback from the LLM itself \cite{shinn2023reflexion}, and 2) Externally provided feedback \cite{peng2023check}.
% LLMs have been extensively employed in various NLP tasks. However, their performance is hindered by issues such as hallucinations and unfaithful reasoning. A proposed solution to these challenges involves incorporating reflection steps \cite{pan2023automatically}. The sources of feedback for reflection are categorized into two types: 1) Self-provided feedback by the LLM \cite{shinn2023reflexion} and 2) Feedback injected through external means \cite{peng2023check}. 
% The first category leverages the LLM itself for both evaluation and refinement, such as SELFCHECK \cite{miao2023selfcheck} and SELF-REFINE \cite{madaan2023self}. It is typically iterative, continuing until the output meets certain criteria or is interrupted in cases of model stagnation. The second category utilizes various external tools to assess and provide feedback on LLM-generated content, such as separately trained models \cite{akyurek2023rl4f}, additional domain-specific knowledge \cite{peng2023check}, and other tools \cite{welleck2022generating}. Feedback through external means offers greater flexibility, introducing information not inherent in LLMs and identifying errors that the LLMs themselves may not detect. 
% The first category focuses on utilizing the LLM for both evaluation and refinement. Techniques such as SELF-CHECK \cite{miao2023selfcheck} and SELF-REFINE \cite{madaan2023self} fall under this category. These methods typically involve an iterative process, where the LLM continuously evaluates and refines its output until certain predefined criteria are met.
The first type uses the LLM for both evaluation and refinement. Techniques like SELF-CHECK \cite{miao2023selfcheck} and SELF-REFINE \cite{madaan2023self} follow an iterative process where the LLM continuously evaluates and refines its output until certain predefined criteria are met.
% The second category involves the use of external tools to provide feedback on the LLM-generated content. These tools can include separately trained models \cite{akyurek2023rl4f}, additional domain-specific knowledge \cite{peng2023check}, and other resources \cite{welleck2022generating}. Feedback from external sources introduces additional flexibility, as it brings in information that may not be inherent in the LLM itself, enabling the identification of errors or inconsistencies that the LLM might overlook. External feedback sources are especially valuable for refining outputs in cases where the LLM’s internal evaluation might fall short.
The second type uses external tools—such as separately trained models \cite{akyurek2023rl4f}, additional domain-specific knowledge \cite{peng2023check}, or other resources \cite{welleck2022generating}—to assess and enhance the LLM’s responses. External feedback adds flexibility by providing information beyond the LLM’s own knowledge, helping to identify errors or inconsistencies the model might miss. It is especially useful when the LLM’s self-evaluation is insufficient.
% In our framework, both intra-modal consistency reflection and inter-modal alignment verification fall under the second category, where a separately trained model helps determine if the LLM needs to reselect candidate entities. The latter also extracts visual clues using image-to-text models and provides them back to the LLM.
In our framework, both intra-modal consistency reflection and inter-modal alignment verification belong to the second category. A separately trained model assesses whether the LLM needs to reselect candidate entities. The latter also extracts visual clues using image-to-text models and fed back into the LLM.

\section{Overview}

In this section, we first formalize the task of multimodal entity linking. Then, we provide an overview of our proposed framework for the task.

\subsection{Task Formulation}
MEL is the task of aligning mentions within multimodal contexts to their respective entities in a KG. Formally, given an input composed of three parts, $\{m, T_m, I_m\}$, where $m$ denotes a mention, $T_m$ is the textual context surrounding $m$, and $I_m$ is the image context for $m$, MEL aims to predict a standard entity $e \in \xi$ for the mention $m$, where $\xi$ is the entity set in the KG.

\subsection{Framework}
% 1. 如图1所示，针对MEL任务的解决方案大致可以分为以下四步：
% 2. Target Entity Selection。我们依次利用字面相关性匹配方法和微调的LLM来为给定Mention从KG中初步筛选出最匹配的一个entity。
% 3. Intra-modal Consistency Reflection。我们通过advanced embedding model计算实体描述和mention的textual context的相似性，以此来衡量步骤1中所选entity和给定mention的语义一致性。
% 4. 如果相似性大于预定义的阈值，则进入下一个步骤。否则，回到步骤一来重新选择新的候选实体。
% 5. Inter-Modal Consistency Verification。我们通过多模态预训练模型来计算实体描述和mention image的对齐程度，以此来衡量所选实体和mention的匹配程度。
% 6. 如果匹配程度大于预定义的阈值，则该entity作为最终输出。否则，进入下一个步骤。
% 7. Visual Iterative Feedback。我们从mention图像中提取视觉线索，并将其作为步骤1的输入，进入下一轮的迭代。
% 8. 需要注意的是，在每次迭代过程中，我们会采用不同的提取策略来捕获mention图像不同方面下的语义信息。

As shown in Figure \ref{fig:framework}, our solution for MEL can be roughly divided into the following four steps: 

\begin{enumerate}
    \item Target Entity Selection (TES). We utilize a lexical relevance matching approach, followed by a fine-tuned LLM, to initially select the Top-1 relevant entity from the KG for a given mention.

    \item Intra-modal Consistency Reflection (ICR). We utilize an advanced embedding model to calculate the similarity between the entity description and the mention's text—both within the text modality—to assess the semantic consistency between the candidate entity selected in Step 1 and the given mention. If the consistency is low, the process returns to Step 1 to select a new candidate entity.
    
    % , checking if the selected entity is semantically consistent with the mention. \textcolor{red}{If the similarity is above a set threshold, we move to the next step. If not, we return to step 1 and reselect a new candidate entity.}

    \item Inter-modal Alignment Verification (IAV). A multimodal pre-trained model is introduced to evaluate the alignment between the entity description and the given mention image—information from different modalities. If the matching degree exceeds the predefined threshold, the entity is selected as the final output. If not, we move to the next step.

    \item Visual Iterative Feedback (VIF). 
    Visual clues are extracted from the mention image and used as the input for step 1 in the subsequent iteration. Note that in each iteration, we use different extraction strategies to capture diverse semantic aspects of the mention image for reasoning.
\end{enumerate}

\section{Methodology}
% 1. 在这一章中，我们将详细阐述我们解决方案中的四个步骤。
In this section, we will provide a detailed explanation of the four steps in our solution.

\subsection{Target Entity Selection}
\label{Target Entity Selection}

\begin{figure}[t]
\centering
\includegraphics[width=0.88\linewidth]{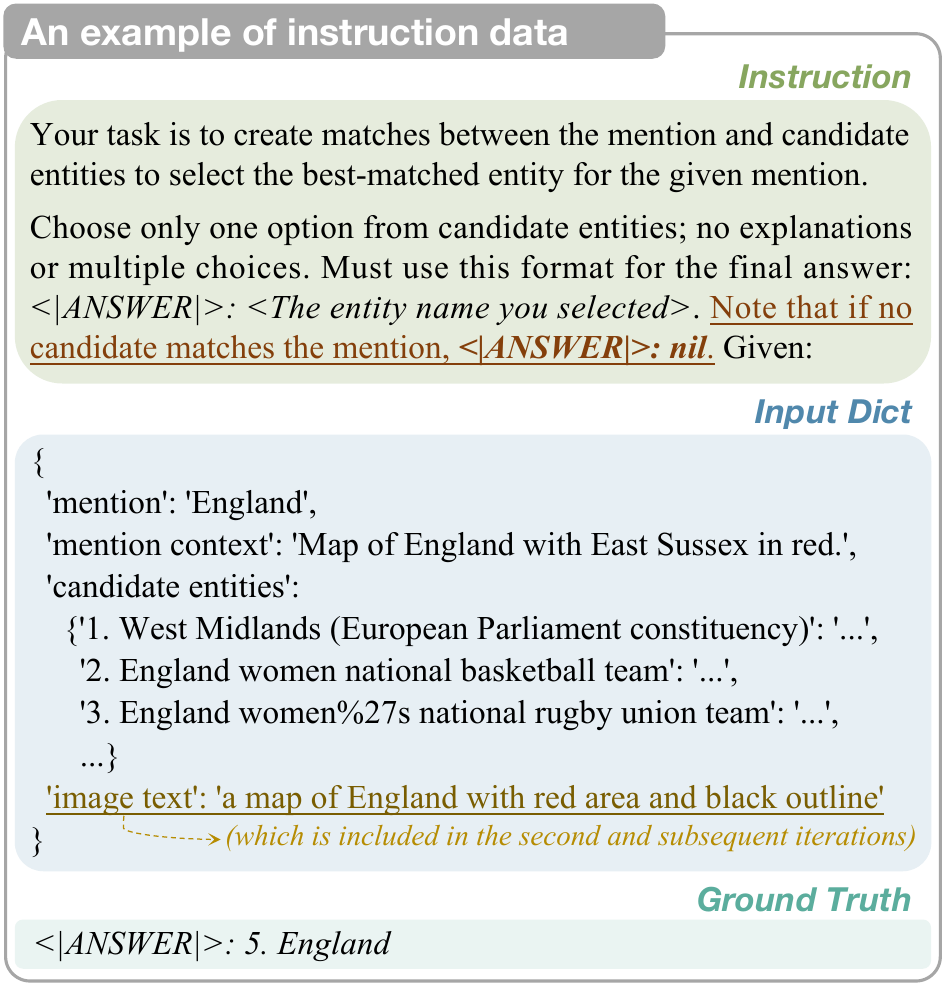}
\caption{Instruction example for target entity selection.}
\label{fig:instruction_data_example}
\end{figure}

The purpose of this step is to select the Top-1 entity from the KG that is most relevant to the given mention. This is achieved in two stages. In the first stage, given a mention $m$ and its surrounding textual context $T_m$, we utilize a fuzzy string matching method\footnote{\url{https://github.com/seatgeek/fuzzywuzzy}} to retrieve the Top-$k$ lexically relevant candidate entities from the KG, along with their descriptions~\cite{wang2022multimodal,wang2022wikidiverse}. This reduces the candidate space and controls the input length for the LLM used in the next stage.
In the second stage, we fine-tune a LLM using an instruction dataset constructed from the training set, where each instance includes: (1) an instruction prompt, (2) a dictionary-style input that includes the mention, its context, and candidate entities, and (3) the ground-truth entity label. An example of this format is shown in Figure~\ref{fig:instruction_data_example}.
The fine-tuned model is then used to select the most appropriate entity $e$, with its description $D_e$, from the Top-$k$ candidates produced in the first stage. 
Two important considerations are addressed in this process: (1) The given mention may not have a corresponding entity in the KG, representing an out-of-KG linking scenario~\cite{shi2024generative}. To handle this, we explicitly instruct the LLM that the output may be empty (i.e., ``nil'') when no suitable entity is found. 2) In the first iteration, the LLM relies only on the mention and its textual context. In the second and subsequent iterations, the input is augmented with visual clues extracted in Step 4 of the previous iteration, enabling the model to make more reasonable selection.

% The purpose of this step is to initially select the Top-1 entity most relevant to the given mention from the KG. 
% Specifically, given a mention $m$ and its textual context $T_m$, we first utilize a fuzzy string matching method\footnote{\url{https://github.com/seatgeek/fuzzywuzzy}} to retrieve the Top-$k$ lexically relevant candidate entities and obtain their descriptions from the KG~\cite{wang2022multimodal,wang2022wikidiverse}. Next, we build an instruction dataset based on the training set and use it to fine-tune an open-source LLM, enabling it to select the entity $e$ (with description $D_e$) that best matches the given mention from the Top-$k$ candidate entities. An example of the instruction data is shown in Figure \ref{fig:instruction_data_example}, which includes instructions, a dictionary converted from the input, and the ground truth in the training set. It should be noted: 1) There may be no corresponding entity for the given mention in the KG, which is an out-of-KG linking problem~\cite{shi2024generative}. To address this, we emphasize in the instruction that the result may be empty (i.e., ``nil''). 2) In the second and subsequent iterations, the data dictionary will also include visual clue information extracted in step 4 from the previous iteration.

\subsection{Intra-modal Consistency Reflection}
\label{Intra-modal Consistency Reflection}
% 1. 这一步的目的是衡量mention和entity是否在文本语义层面上具有一致性。
% 2. 具体而言，我们首先将mention m、mention textual context Tm进行拼接，得到Cm。
% 3. 如果当前迭代的轮次大于等于2，则还需在拼接的过程加上mention的image text I'm（具体的获取方式见【Visual Iterative Feedback】）。
% 4. 其次，我们将entity e和its 描述De进行拼接，得到Ce。
% 5. 最终，我们利用一个embedding model来分别获得Cm和Ce的语义向量，并采用点积with归一化来计算这两个向量的余弦相似性。
% 6. 
% 7. 上述过程可以形式化地表示为：
This step aims to evaluate the semantic consistency between the given mention and the candidate entity within the textual modality. To construct a comprehensive textual representation of the mention, we concatenate the mention $m$ with its textual context $T_m$ to form $C_m$. If the current iteration is at least 2, we additionally incorporate the mention’s image-derived text $I'm$ (see Section \ref{Visual Iterative Feedback} for details), resulting in a richer and more informative input. Formally, this process is defined as:
\begin{equation}
    C_m=\left\{
    \begin{aligned}
        Concate(m, T_m), \ \ \ \ \ r=1 \\
        Concate(m, T_m, I'm), \ \ \ \ \ r \geq 2
    \end{aligned}
    \right.
\end{equation}
Similarly, we construct a textual representation for the entity by concatenating the entity name $e$ with its associated description $D_e$, selected during Step 1, i.e., $C_e = Concate(e, D_e)$. Then, we employ an advanced embedding model to encode both $C_m$ and $C_e$ into semantic vector representations. The degree of semantic alignment is measured by computing the similarity between the two embeddings, implemented as a normalized dot product:
\begin{equation}
    Score_{icr}=Norm\left(Embed(C_m) \cdot Embed(C_e)\right),
\end{equation}
where $Norm(\cdot)$ represents the normalization operation. If $Score\_{icr}>\alpha$, where $\alpha$ is a predefined threshold, the candidate entity $e$ is considered semantically consistent with the mention $m$. Otherwise, the process returns to the step 1, removes $e$ from the candidate set, and proceeds to reselect a more appropriate entity.
% 分段函数
% Cm=Concanate（m，Tm） r=1
% Cm=Concanate（m，Tm, I'm）r>=2
% Ce=Concanate（e, De）
% Score\_icr=Norm(Embed(Cm), Embed(Ce))
% 8. 其中，Norm表示归一化操作。若Score\_icr>XX，其中XX是一个预定义的阈值，则说明entity e与mention m在语义上具有一致性的。否则，返回第一步，并从候选项中排除entity e后再重新选择目标实体。
This process continues until the selected entity is semantically consistent with the given mention or the predefined iteration limit is reached. If the limit is reached without finding a reasonable entity, the last selected entity is used as the target.
% 1. 这一步骤将迭代进行，直到选定的实体与给定mention具有语义一致性，或者达到预定义的迭代次数限制。在后一种情况下，选择最后一次选择的实体作为输出。

\subsection{Inter-Modal Alignment Verification}
\label{Inter-Modal Alignment Verification}
% 1. 这一步的目的是利用上一步选中实体和给定mention的不同模态信息来衡量他们的对齐程度。
% 2. 具体而言，给定实体e的描述De和mention image Im，我们使用多模态预训练模型（如CLIP）来分别获得它们的文本向量和图像向量，并用一个点积操作计算上述两个向量之间的余弦相似性。
% 3. 这一过程可以形式化被表示为：

The objective of this step is to verify whether the entity selected in the previous step remains semantically consistent with the given mention across different modalities. 
Specifically, given the textual description of the candidate entity $D_e$ and the mention image $I_m$, we utilize a multimodal pre-trained model (e.g., CLIP~\cite{radford2021learning}) to project both inputs into a shared embedding space. The model's text encoder $Enc_T$ processes the entity description $D_e$, while the image encoder $Enc_I$ encodes the mention image $I_m$. The dot product between the resulting embeddings yields a similarity score that reflects the cross-modal alignment between the entity and the mention. This can be formally defined as:
\begin{equation}
    Score_{iav}=Enc_T(D_e) \cdot Enc_I(I_m),
\end{equation}
where $Score_{iav}$ denotes the inter-modal alignment score. If this score exceeds a predefined threshold $\beta$, the candidate entity $e$ is taken as the final result. Otherwise, the framework proceeds to the next step to explore alternative candidates using additional multi-modal information.

% $Enc_T$ and $Enc_I$ denote the text and image encoders of the multimodal pre-trained model, respectively. If $Score\_iav>\beta$, where $\beta$ is a predefined threshold, entity $e$ is taken as the final result. Otherwise, the process continues to the next step.
% Score\_iav=Enc\_T(De)点乘Enc\_I(Im)
% 4. 其中，Enc\_T和Enc\_I分别表示多模态预训练模型中的文本和图像编码器。若Score\_iav>YY，其中YY是一个预定义的阈值，则将entity e作为最终结果，否则，我们将继续进入到下一步。

\subsection{Visual Iterative Feedback}
\label{Visual Iterative Feedback}
% multi-faceted descriptions
% 1. 这一步的目的是挖掘mention image中的视觉线索，并将该线索反馈到step 1中，以此来辅助仅靠textual context不一定能精准识别目标entity的模型
% 2. 在这里，我们采用多个不同的image-to-text models（如“OCR”, “Image Captioning”, “Dense Captioning”, and “Image Tagging”）来为mention image生成不同方面的文本描述I‘m。
% 3. 为了防止信息过载，我们在每次迭代的过程中，仅采用一个model来生成文本。
% 4. 这些models的使用顺序通过验证集来确定。
% 5. 如果所有models使用完，模态间对齐验证仍没有通过，那我们将选择第一步选择的那个top-1实体。
The purpose of this step is to extract visual clues from the mention image and provide them as additional input to Step 1, aiding the model in cases where textual context alone is insufficient for accurate entity identification. To achieve this, we leverage multiple image-to-text models (e.g., OCR, Image Captioning, Dense Captioning, and Image Tagging) to generate a variety of textual descriptions $I'm$ for the mention image $I_m$. To prevent information overload, only one model is used per iteration, and the resulting text description is incorporated into the input data dictionary of the LLM in Step 1. The order in which the models are applied is determined based on the validation set to ensure optimal performance.  If all models have been exhausted and inter-modal alignment still fails, the entity selected in Step 1 with the highest score is retained as the final result.

% Here, we utilize multiple image-to-text models (e.g., ``OCR'', ``Image Captioning'', ``Dense Captioning'', and ``Image Taging'') to generate different text descriptions $I'm$ for the mention image $I_m$. To avoid information overload, only one model is used per iteration, and the produced text description is added to the input data dictionary of the LLM in the first step. The order of model usage is determined by the validation set. If all models have been applied and inter-modal alignment still fails, the Top-1 entity selected in Step 1 is chosen as the result.

\section{Experiments}
In this section, we conduct extensive experiments to evaluate our proposed framework and provide detailed analyses to highlight its effectiveness.

% 此外，我们提供了针对我们方法的详细分析来进一步说明它的有效性。

\subsection{Experimental Setup}
\textbf{Datasets.}
% 1. In this study，我们使用三个数据集进行评估，即WikiMEL[1]、WikiDiverse[2]和Richpedia[3]。
% 2. WikiMEL collects data from Wikipedia’s entity pages, containing more than 22k multimodal sentences, with its primary entity type being Person. It uses Wikidata as its target KG.
% 3. WikiDiverse is built by Wikinews, covering 7 types of entities(i.e., Person, Organization, Location, Country, Event, Works, and Misc). It utilizes Wikipedia as its target KG.
% 4. Richpedia collects the Wikidata index of entities in a largescale multimodal knowledge graph Richpedia, and is ob.tained the multimodal information from Wikipedia. (改写)
% 5. 我们遵循原始的数据集划分。在WikiDiverse中，训练集、验证集、测试集的比例为8:1:1。对于WikiMEL和Richpedia，划分的比例为7:1:2.
% 6.The statistics of three datasets are concluded in Table 1. 
% 缺张表
We employ three widely used datasets for evaluation: WikiMEL~\cite{wang2022multimodal}, WikiDiverse~\cite{wang2022wikidiverse}, and Richpedia-MEL (denoted as RichMEL)~\cite{wang2022multimodal}. WikiMEL collects data from Wikipedia’s entity pages, containing more than 22k multimodal sentences, with its primary entity type being Person. It uses Wikidata as its target KG. WikiDiverse is built by Wikinews, covering 7 types of entities (i.e., Person, Organization, Location, Country, Event, Works, and Misc). It utilizes Wikipedia as its target KG. RichMEL compiles the Wikidata information for entities in Richpedia~\cite{wang2020richpedia} and gathers multimodal data from Wikipedia. We adhere to the original dataset splits: WikiDiverse follows an 8:1:1 partition for training, validation, and test sets, respectively, while WikiMEL and RichMEL utilize a 7:1:2 split. Dataset statistics are summarized in Table \ref{tab:statistics}.

\begin{table}[t]
\centering
\caption{Statistics of three MEL datasets. }
\scalebox{0.9}{
\begin{tabular}{>{\centering\arraybackslash}p{3cm} >{\centering\arraybackslash}p{1.6cm} >{\centering\arraybackslash}p{1.6cm} >{\centering\arraybackslash}p{1.6cm}}
\toprule
             & WikiMEL & Wikidiverse & RichMEL \\ \midrule
\# samples    & 22,136   & 7,824        & 17,806     \\
\# mentions           & 25,846   & 16,327       & 18,752     \\
\# entities           & 17,890   & 78,556       & 72,085     \\
\# mention images & 22,136   & 12,909       & 15,853     \\
\# text length        & 8.2     & 10.2        & 13.6      \\ \bottomrule
\end{tabular}
}
\label{tab:statistics}
\end{table}

\textbf{Baselines.}
We compare our proposed framework with various SoTA methods,
% \footnote{We choose not to compare with the recent work, Unimel~\cite{liu2024unimel}, for two main reasons: (1) Unimel cannot be reproduced because the authors did not provide a fine-tuned training set or model checkpoint; and (2) the method prunes the original data, making it unable to handle out-of-KG linking problem.}, 
which are divided into two groups: 
\begin{itemize}
    \item Text-only methods:
    BERT~\cite{kenton2019bert} is used to encode the mention and entity to extract features for relevance scoring. RoBERTa~\cite{liu2019roberta} is a BERT variant. BLINK~\cite{wu2020scalable} is a two-stage method: retrieval and re-ranking.
    
    % , including BERT~\cite{kenton2019bert}, RoBERTa~\cite{liu2019roberta}, and BLINK~\cite{wu2020scalable}.

    \item Visual-text fusion methods: CLIP~\cite{radford2021learning} uses Transformer-based encoders for visual-text feature extraction. DZMNED~\cite{moon2018multimodal} combines visual, textual, and character features with modality attention. JMEL~\cite{adjali2020multimodal} maps visual and textual features to a joint space. GHMFC~\cite{wang2022multimodal} employs gated fusion and contrastive training for entity linking. MIMIC~\cite{luo2023multi} fuses features via interaction units with contrastive learning. DRIN~\cite{xing2023drin} uses a dynamic GCN for mention-entity alignment. GEMEL~\cite{shi2024generative} leverages LLMs for generative linking with in-context learning. DWE~\cite{song2024dual} integrates text and visual features, refining with Wikipedia. DWE+\cite{song2024dwe+} is a vriant of DWE. OT-MEL~\cite{zhang2024optimal} formulates the correlation assignment as an optimal transport problem. Unimel~\cite{liu2024unimel} use LLMs to select the targe entity.

    % , which include CLIP~\cite{radford2021learning}, DZMNED~\cite{moon2018multimodal}, JMEL~\cite{adjali2020multimodal}, GHMFC~\cite{wang2022multimodal}, MIMIC~\cite{luo2023multi}, DRIN~\cite{xing2023drin}, GEMEL~\cite{shi2024generative}, DWE~\cite{song2024dual}, DWE+~\cite{song2024dwe+}, OT-MEL~\cite{zhang2024optimal}, and Unimel~\cite{liu2024unimel}.
\end{itemize}
% Detailed descriptions of these baselines are provided in Appendix \ref{app:baselines}.

% 1. 我们没有对比最新的工作Unimel[18]，有两个原因：
% 2. 1）Unimel无法复现，因为他们没有提供微调的训练集和微调后模型的checkpoint。
% 3. 2）该工作对原始数据进行了删减，无法处理out-of-KB linking problem。

\textbf{Metrics.}
Following previous works~\cite{song2024dwe+,zhang2024optimal}, we employ the Top-$K$ accuracy metric for evaluation:
\begin{equation}
    Top\text{-}K = \frac{1}{N}\sum_{i=1}^N I(g_i \in P_i^K),
\end{equation}
where $N$ denotes the total number of samples, $g_i$ is the ground truth entity, $P_i^K$ represents the set of the top-$K$ predicted entities, and $I(\cdot)$ is the indicator function.
The process of selecting the Top-K ($K$=3 or 5) results in our I2CR framework is as follows: In the first step (TES), we iteratively perform the process until the number of entities in the candidate set reaches $K$. In the second (ICR) and third (IAV) steps, we rank the entities that do not meet the judgment criteria at the end of the set. If none of the entities in the set meet the criteria, we iteratively reselect Top-$K$ candidate entities.

% 不同方法在三个数据集上的对比结果。加粗表示最好的结果，下划线表示第二好的结果。右下角括号中的数字表示三轮结果的标准差。
\begin{table*}[t]
\centering
\caption{Comparison results of different methods on three datasets. Bold indicates the best result, and underline indicates the second best. The numbers in parentheses represent the standard deviation from three rounds of results.}
\scalebox{0.9}{
\begin{tabular}{@{}>{\centering\arraybackslash}p{1.9cm}|>{\centering\arraybackslash}p{1.9cm}|>{\centering\arraybackslash}p{1.35cm}>{\centering\arraybackslash}p{1.35cm}>{\centering\arraybackslash}p{1.35cm}|>{\centering\arraybackslash}p{1.35cm}>{\centering\arraybackslash}p{1.35cm}>{\centering\arraybackslash}p{1.35cm}|>{\centering\arraybackslash}p{1.35cm}>{\centering\arraybackslash}p{1.35cm}>{\centering\arraybackslash}p{1.35cm}@{}}
\toprule
\multirow{2}{*}{\textbf{Modality}} & \multirow{2}{*}{\textbf{Model}} & \multicolumn{3}{c|}{\textbf{WikiMEL}} & \multicolumn{3}{c|}{\textbf{Wikidiverse}} & \multicolumn{3}{c}{\textbf{RichMEL}} \\ \cmidrule(l){3-11} 
                                   &                                 & Top-1       & Top-3      & Top-5      & Top-1        & Top-3        & Top-5       & Top-1       & Top-3       & Top-5      \\ \midrule
\multirow{3}{*}{Text-only}         & BERT                            & 31.7        & 40.4       & 48.8       & 22.2         & 37.9         & 53.8        & 31.6        & 36.1        & 42.0         \\
                                   & BLINK                           & 30.8        & 38.9       & 44.6       & 70.9         & 72.8         & 74.2        & 30.8        & 35.7        & 38.8       \\
                                   & RoBERTa                         & 73.8        & 85.9       & 89.8       & 59.5         & 78.5         & 85.1        & 61.3        & 81.6        & 87.2       \\ \midrule
\multirow{10}{*}{Visual-text}      & CLIP                            & 83.2        & 92.1       & 94.5       & 61.2         & 79.6         & 85.2        & 67.8        & 85.2        & 90.0         \\
                                   & DZMNED                          & 78.8        & 90.0         & 92.6       & 56.9         & 75.3         & 81.4        & 68.2        & 82.9        & 87.3       \\
                                   & JMEL                            & 64.7        & 80.0         & 84.3       & 37.4         & 54.2         & 61.0          & 48.8        & 66.8        & 74.0         \\
                                   & GHMFC                           & 76.6        & 88.4       & 92.0         & 60.3         & 79.4         & 84.7        & 72.9        & 86.9        & 90.6       \\
                                   & MIMIC                           & 88.0          & 95.1       & 96.4       & 63.5         & 81.0           & 86.4        & 81.0          & 91.8        & 94.4       \\
                                   & DRIN                            & 65.5        & -          & 91.3       & 51.1         & 77.9         & 89.3        & -           & -           & -          \\
                                   & DWE                             & 44.5        & 59.7       & 66.3       & 46.6         & 74.5         & 80.9        & 64.6        & 87.4        & 96.6       \\
                                   & DWE+                            & 44.9        & 60.4       & 67.2       & 47.1         & 75.6         & 81.3        & 67.0          & 88.3        & \underline{97.1}       \\
                                   & GEMEL                           & 82.6        & -          & -          & 86.3         & -            & -           & -           & -           & -          \\               
                                   & OT-MEL                          & \underline{89.0}          & \underline{95.6}       & \underline{97.0}         & 66.1         & 82.8         & 87.4        & 83.3        & 92.4        & 94.8       \\

                                   & UniMEL                          & 88.7          & 95.2       & 96.9         & \underline{86.5}         & \underline{89.8}         & \underline{90.4}        & \underline{85.2}        & \underline{92.6}        & 96.8       \\
                                   
                                   \midrule
Visual-text                              & I2CR(ours)                            & \textbf{92.2}\footnotesize{(0.24)}        & \textbf{96.1}\footnotesize{(0.17)}       & \textbf{97.5}\footnotesize{(0.05)}       & \textbf{91.6}\footnotesize{(0.26)}         & \textbf{94.7}\footnotesize{(0.22)}         & \textbf{95.6}\footnotesize{(0.08)}        & \textbf{86.8}\footnotesize{(0.20)}        & \textbf{92.9}\footnotesize{(0.16)}        & \textbf{97.2}\footnotesize{(0.05)}       \\ \bottomrule
\end{tabular}
}
\label{tab:main}
\end{table*}

\textbf{Implementation details.}
% 1. In 3.1, we set k = 10.
% 2. 我们使用LLaMA3-8B[19]作为基座LLM，并使用LoRA [25]方法对该模型进行微调with AdamW优化器，学习率为2e-5，batch\_size 大小为1，梯度累积步数为16，预热步数为100。
% 3. 在推理过程中，温度设置为0.9，其他参数保持在其默认设置。
During training, we fine-tune Llama3-8B~\cite{dubey2024llama} using LoRA~\cite{hu2021lora} with the AdamW optimizer, a learning rate of 2e-5, batch size 1, gradient accumulation steps of 16, and 100 warm-up steps. 
To test the robustness and generalization of our framework, we train it only on WikiDiverse and test it on three datasets. Since the entity descriptions in WikiDiverse are longer and more complex, using them directly as input may exceed the LLM's input limits and cause the model to truncate the text. This would affect the model's ability to extract and understand the information accurately. To solve this, we follow prior work~\cite{liu2024unimel} and use Llama3-8B with a prompt to summarize the descriptions, shortening them to a length similar to those in WikiMEL and RichMEL.
For inference, the temperature is set to 0.9, with other parameters at defaults.
In Section \ref{Target Entity Selection}, we set $k$=10.
In Section \ref{Intra-modal Consistency Reflection}, we consider SFR-Embedding-Mistral~\cite{meng2024sfrembedding} as our embedding model. We set $\alpha$=0.5 for WikiMEL, 0.8 for WikiDiverse, and 0.75 for RichMEL.
In Section \ref{Inter-Modal Alignment Verification}, we use CLIP\footnote{CLIP-ViT-bigG-14-laion2B-39B-b160k} as the multimodal pre-trained model and set $\beta$=31 for all three datasets.
$\alpha$ and $\beta$ are determined through the validation set.
In Section \ref{Visual Iterative Feedback}, we employ four latest models (i.e., ``OCR'', ``Image Captioning'', ``Dense Captioning'', and ``Image Tagging'') from Azure Cognitive Services APIs\footnote{\url{https://portal.azure.com/\#view/Microsoft\_Azure\_ProjectOxford/CognitiveServicesHub//ComputerVision}} as the image-to-text models.
Our experiments are conducted on a workstation running Ubuntu 20.04.6 LTS, with two NVIDIA Tesla A100 GPUs, and 80GB of memory.

% 1. In section 3.3，我们将CLIP【脚注CLIP-ViT-bigG-14-laion2B-39B-b160k】作为多模态预训练模型，并set the φ to 31 for three datasets.

% 1. 上述θ 和 φ 的值是通过验证集确定的。

% 1. 在Section 3.4中，我们采用the latest models from Azure Cognitive Services APIs[24]。

\subsection{Main Results}
% 1. 我们在WIKIMEL, WikiDiverse and RichMEL三个数据集上对比了我们的I2CR方法和多个具有竞争力的baselines。
% 2. 在这三个数据集上的Top-1、Top-3、Top-5的准确性如表2所示。
We evaluate our framework against several strong baselines on three datasets: WIKIMEL, WikiDiverse, and RichMEL. The top-1, top-3, and top-5 accuracies for all models on these datasets are presented in Table \ref{tab:main}.

% 1. From the results, we observe that 与现有的state-of-the-art (SOTA) 方法相比，我们的I2CR在三个数据集上均取得了最好的性能，说明了其有效性。
% 2. 具体而言，我们的方法在三个数据集上分别获得了92.2\%、91.6\%、86.8\%的Top-1准确率，相较于之前最好的方法分别提升了3.2\%、5.3\%、3.5\%。
% 2. 此外， 相较于WikiMel和Wikidiverse数据集，模型在RichMEL上的性能相对于低一点。
% 3. 根据我们对错误案例的观察，我们发现RichMEL中包含需要强先验知识的样本。
% 4. 比如说，Mention是一个运动员的名字，但是textual context是这个mention出演过的电影经历，图像是一张证件照。
% 5. 这要求模型需要具备这位运动员曾经有电影经历的先验知识。
From the results, we observe that our I2CR framework outperforms existing SoTA methods on all three datasets, demonstrating its effectiveness. Specifically, our framework achieves top-1 accuracies of 92.2\%, 91.6\%, and 86.8\% on the three datasets, respectively, which are 3.2\%, 5.1\%, and 1.6\% higher than the previous best methods. Notably, our framework is trained only on WikiDiverse, not on WikiMEL or RichMEL, yet it still achieves SoTA performance on these two datasets. This demonstrates the robustness and strong generalization ability of our framework.
In addition, the performance on RichMEL is relatively lower compared to WikiMEL and WikiDiverse. From our analysis of the error cases, we find that RichMEL contains samples that require more strong prior knowledge and complex reasoning. For instance, ``mention'' refers to an athlete, but the textual context describes this individual’s movie experience, and the accompanying image is an ID photo. This type of samples requires the model to possess prior knowledge that the athlete has also been involved in movies, which poses a greater challenge.

% "b"、"c"、"d"为框架图中各模块的序号，分别表示ICR, IAV, and VIF模块。
\begin{table}[t]
\centering
\caption{Ablation study of overall framework. ``b'', ``c'', and ``d'' correspond to the module labels in the framework diagram, representing the ICR, IAV, and VIF modules, respectively.}
\scalebox{0.9}{
\begin{tabular}{>{\centering\arraybackslash}p{2.4cm} >{\centering\arraybackslash}p{1.8cm} >{\centering\arraybackslash}p{1.8cm} >{\centering\arraybackslash}p{1.8cm}}
\toprule
\multirow{2}{*}{Model} & \multicolumn{3}{c}{Top-1 Accuracy (\%)} \\ \cmidrule(l){2-4} 
                       & WikiMEL   & Wikidiverse   & RichMEL   \\ \midrule
I2CR                   & 92.2      & 91.6          & 86.8        \\ \midrule
w/o b                  & 91.2      & 90.1          & 85.7        \\
w/o c                  & 90.1      & 90.7          & 85.6        \\
w/o d                  & 88.0        & 89.4          & 84.7        \\ \midrule
w/o bc                 & 89.3      & 89.3          & 85.0          \\
w/o bd                 & 87.3      & 89.1          & 84.4        \\
w/o cd                 & 86.4      & 88.9          & 83.7        \\ \midrule
w/o bcd                & 86.0        & 88.6          & 83.3        \\ \bottomrule
\end{tabular}}
\label{tab:ablation1}
\end{table}

\subsection{Ablation Study}
% 1. 整体框架中模块的消融
% 1. 为了验证我们框架中每一模块的有效性，我们在保留TES模块的基础上，枚举剩余三个模块(i.e., ICR、IAV以及VIF)所有的删除组合。
% 2. 结果如表3所示。
% 3. 从图中，我们观察到：
% 4. 1）删除任何一个模块都会导致模型性能的下降，下降了至少1\%或以上。
% 5. 这证明了我们的框架中所有模块的有效性。
% 6. 2）与单独删除ICR或IAV相比，单独删除VIF之后性能下降最大，说明了视觉clue对MEL任务的重要性。
% 7. 删除两个模块的三组实验也证明了这一结论。
% 8. 3）如果ICR、IAV以及VIF三个模块都删除，即保留仅用文本的TES模块，模型也能达到较好的效果。
% 9. 这说明了TES模块有效性地挖掘了文本信息蕴含的MEL的主导线索。
\textbf{Ablation of modules in the overall framework.} 
To evaluate the effectiveness of each module in our framework, we retain the TES module and systematically delete combinations of the remaining three modules (i.e., ICR, IAV, and VIF). The results are in Table \ref{tab:ablation1}. From the results, we notice the following: 1) Deleting any module leads to a decrease in performance, with a drop of at least 1\% or more, validating the contribution of each module in our framework. 2) Deleting VIF alone leads to the largest decrease in performance compared to deleting ICR or IAV, highlighting the crucial role of visual information for our framework. This finding is further confirmed by experiments where two modules are deleted simultaneously. 3) Even when all three modules (ICR, IAV, and VIF) are deleted, leaving only the TES module with textual input, the model still achieves relatively good performance. This suggests that the TES module is effective in extracting the key clues from text alone.

\begin{table}[t]
\centering
\caption{Ablation study of VIF module. ``ocr'', ``cap'', ``den'', ``tag'' indicate ``OCR'', ``Image Captioning'', ``Dense Captioning'', and ``Image Tagging'' models, respectively.}
\scalebox{0.9}{
\begin{tabular}{>{\centering\arraybackslash}p{2.8cm} >{\centering\arraybackslash}p{1.7cm} >{\centering\arraybackslash}p{1.7cm} >{\centering\arraybackslash}p{1.7cm}}
\toprule
\multirow{2}{*}{Model} & \multicolumn{3}{c}{Top-1 Accuracy (\%)} \\ \cmidrule(l){2-4} 
                       & WikiMEL   & Wikidiverse   & RichMEL   \\ \midrule
I2CR                   & 92.2      & 91.6          & 86.8        \\ \midrule
w/o ocr                & 91.7      & 91.1          & 86.0          \\
w/o cap                & 91.3      & 90.7          & 86.3        \\
w/o den                & 91.9      & 91.3          & 86.5        \\
w/o tag                & 91.8      & 91.3          & 86.6        \\ \midrule
w/o ocr, cap           & 90.4      & 90.0          & 85.4        \\
w/o ocr, den           & 91.4      & 90.5          & 85.7        \\
w/o ocr, tag           & 91.1      & 90.7          & 85.9        \\
w/o cap, den           & 91.3      & 90.2          & 86.0          \\
w/o cap, tag           & 90.8      & 90.6          & 86.1        \\
w/o den, tag           & 91.5      & 90.9          & 86.3        \\ \midrule
w/o ocr, cap, den      & 89.3      & 89.6          & 85.0          \\
w/o ocr, cap, tag      & 88.7      & 89.7          & 85.2        \\
w/o ocr, den, tag      & 90.2      & 90.1          & 85.4        \\
w/o cap, den, tag      & 89.8      & 90.0          & 85.7        \\ \midrule
w/o all                & 88.0      & 89.4          & 84.7        \\ \bottomrule
\end{tabular}}
\label{tab:ablation2}
\end{table}

% 2. VIF中子模块的消融
% 1. 为了验证VIF中不同faceted视觉线索的有效性，我们枚举了4个image-to-text models是否使用的所有组合。
% 2. 结果如表3所示。
% 3. 从图中，我们观察到：
% 4. 1）VIF模块中的四个子模块对整个框架的迭代过程都有积极的贡献，从而说明了所有侧面的视觉线索都是有用的。
% 5. 2）对于不同的数据集，每个子模块带来的增益大小是有差异的。
% 6. 比如对于WikiMEL和WikiDiverse而言，image caption带来的增益是最大的，而对于RichMEL来说，OCR text更为重要。
\textbf{Ablation of sub-modules in VIF.} 
To evaluate the effectiveness of different faceted visual clues in VIF, we systematically test all possible combinations of the four image-to-text models. The results are shown in Table \ref{tab:ablation2}. From the results, we conclude that: 1) All four sub-modules in VIF contribute positively to the overall performance of the framework, demonstrating that each faceted visual clue is valuable. 2) The contribution of each sub-module varies depending on the dataset. For instance, on WikiMEL and WikiDiverse, the ``image captioning'' sub-module provides the most significant improvement, while on RichMEL, ``OCR'' proves to be more important. In this study, we integrate all four sub-modules into the framework.

\subsection{Detailed Analysis}
% 训
% 我们框架=训+我们流程
% Robustness to other LLMs
% 1. 为了进一步分析我们方法的通用性，我们在不同类型的LLMs上进行了适配，包括open-sourced LLMs（Qwen 2.5-7B、Vicuna1.5-7B、Llama3-8B、Llama3-13B）和close-sourced LLMs（gpt-3.5-turbo和gpt-4o）.
% 2. 针对每个开源模型，我们对比仅用训练集微调的LLM和用我们框架的LLM。
% 3. 针对每个闭源模型，我们对比直接推理的LLM和用我们框架推理的LLM。
% 4. 需要注意的是，由于闭源模型涉及API调用的成本问题，我们分别从三个数据集中随机采样100条数据进行实验。
% 5. 实验结果如表5所示。
\textbf{Versatility of our I2CR framework across different LLMs.}
To further assess the versatility of our framework, we apply it to various LLMs, including both open-source models (Qwen 2.5-7B~\cite{yang2024qwen2}, Vicuna1.5-7B\footnote{\url{https://huggingface.co/lmsys/vicuna-7b-v1.5}}, Llama3-8B, Llama3-13B~\cite{dubey2024llama}) and closed-source models (GPT-3.5-turbo and GPT-4o). For open-source models, there are two options: fine-tuning the LLM on the training set directly, or fine-tuning it using our framework. For closed-source models, the options are: using the LLM directly for inference, or using the LLM for inference with our I2CR framework.
Since using closed-source models incurs API costs, we randomly select 100 samples from each of the three datasets for experiments. The results are shown in Table \ref{tab:analysis1}. Based on the results, we conclude that our framework improves performance for both open-source and closed-source LLMs, even for the larger 13B LLMs.

% 1. 从结果中，我们总结我们的方法不管是适配到开源LLM还是闭源LLM都能带来效果的提升。
% 3. 此外，当我们的模型适配到13B模型的时候，这个结论也是同样有效的。

% 我们方法适配到不同模型的结果。每个LLM有两行结果，第一行表示第一种实验设置，第二行表示第二种实验设置。
% {>{\centering\arraybackslash}p{2.5cm} >{\centering\arraybackslash}p{1.6cm} >{\centering\arraybackslash}p{1.6cm} >{\centering\arraybackslash}p{1.6cm}}
\begin{table}[t]
\centering
\caption{Results of adapting our framework to different LLMs. Each LLM has two rows: The first row represents the output of the base LLM, while the second row represents the output of the LLM with our I2CR framework.}
\scalebox{0.9}{
\begin{tabular}{p{2.8cm} >{\centering\arraybackslash}p{1.7cm} >{\centering\arraybackslash}p{1.7cm} >{\centering\arraybackslash}p{1.7cm}}
\toprule
\multirow{2}{*}{Model} & \multicolumn{3}{c}{Top-1 Accuracy (\%)} \\ \cmidrule(l){2-4} 
                       & WikiMEL   & Wikidiverse   & RichMEL   \\ \midrule
Qwen 2.5-7B            & 88.4      & 87.5          & 79.8        \\
Qwen 2.5-7B++           & 90.6      & 89.1          & 82.1        \\
Vicuna1.5-7B           & 51.3      & 72.0          & 62.9        \\
Vicuna1.5-7B++          & 66.7      & 74.7          & 69.0        \\
Llama3-8B              & 86.0      & 88.6          & 83.3        \\
Llama3-8B++             & 92.2      & 91.6          & 86.8        \\
Llama3-13B             & 87.7      & 88.7          & 82.5        \\
Llama3-13B++            & 92.6      & 91.3          & 85.9        \\ \midrule
GPT-3.5-turbo          & 89.0      & 62.0          & 84.0        \\
GPT-3.5-turbo++         & 91.0      & 74.0          & 86.0        \\
GPT-4o                 & 93.0      & 68.0          & 85.0        \\
GPT-4o++                & 97.0      & 81.0          & 88.0        \\ \bottomrule
\end{tabular}}
\label{tab:analysis1}
\end{table}
\begin{figure}[t]
\centering
\includegraphics[width=0.98\linewidth]{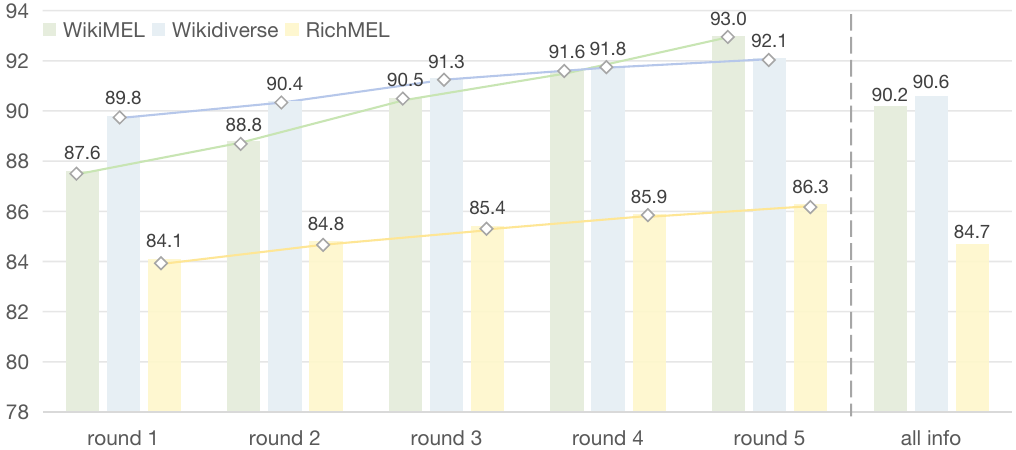}
\caption{Results across different iteration rounds and a single iteration using all visual clues.}
\label{fig:round_all}
\end{figure}

\begin{figure}[t]
\centering
\includegraphics[width=0.92\linewidth]{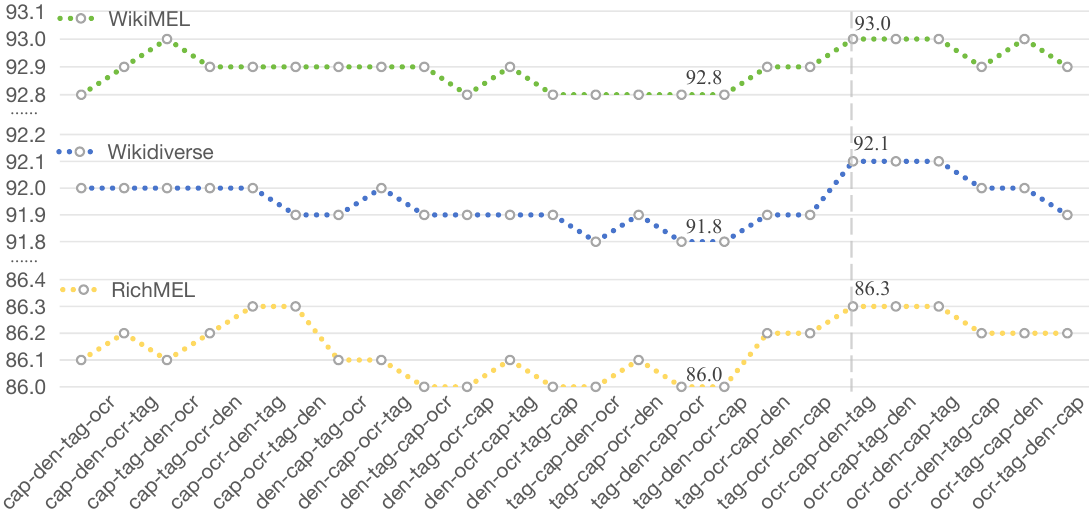}
\caption{Results of using visual clues in different orders.}
\label{fig:sort_all}
\end{figure}

\begin{figure*}[t]
\centering
\includegraphics[width=0.94\linewidth]{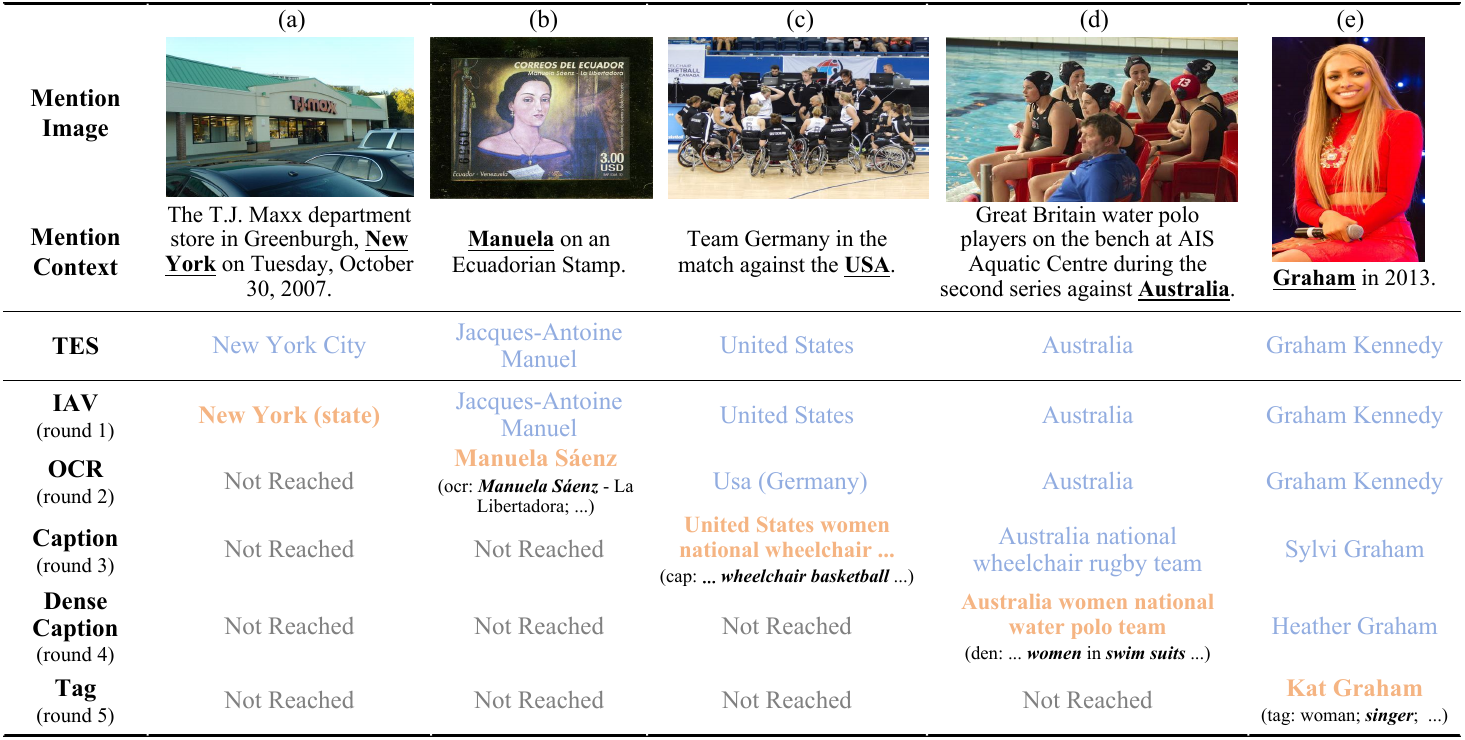}
\caption{Case study. Blue for incorrect output, orange for correct output. ``Not Reached'' indicates that the corresponding sample already has a final answer prior to this round.}
\label{fig:case}
\end{figure*}

% 不同视觉线索给模型性能带来增益的分析
% 1. 为了分析在整个迭代过程中视觉线索的有效性，我们提供了模型在不同迭代轮次（第0轮、第1轮。。。第4轮）的Top-1准确率。
% 2. 此外，我们还比较了只进行一次迭代时的效果，在此过程中，我们将所有视觉线索整合在一起反馈给第一个步骤。
% 3. 实验结果如图3所示。
% 4. 从结果中，我们发现
% 5. 1）随着迭代轮次的增加，模型在三个数据集上的Top-1准确率也会逐渐提升。
% 6. 这说明了不同的视觉线索对模型都是有增益的。
% 7. 2）一次性输入所有的视觉线索相比于第0次的效果有微弱的提升，但是相比于每次仅输入单一视觉线索的效果要差的多。
% 8. 这说明了多个视觉线索的集成可能会导致信息过载，从而导致模型难以精准捕捉到其中丰富的关键信息。
\textbf{Impact of visual clues on model performance.} 
To assess the effectiveness of visual clues throughout the iteration process, we report the top-1 accuracy of the model at the different iteration round (round 1, round 2, ..., round 5). Additionally, we compare the effect of using a single iteration, where all visual clues are integrated and fed back into the first step. The experimental results are presented in Figure \ref{fig:round_all}. From these results, we observe: 1) As the number of iteration rounds increases, the model's top-1 accuracy on the three datasets gradually improves, indicating that the different visual clues are beneficial to the model. 2) Feeding all visual clues at once shows a slight improvement over the 1st round, but performs much worse than using a single visual clue per iteration. This suggests that the integration of multiple clues may lead to information overload, making it harder for the model to accurately capture the key information. 

% 视觉线索使用顺序的分析
% 1. 在VIF模块中，我们在每次迭代过程中不重复地从4个image-to-text models中选择一个来提取mention image的视觉线索。
% 2. 在这个实验中，我们分析视觉线索使用的顺序对模型性能的影响。
% 3. 实验结果如图2所示。
% 4. 从结果中，我们发现不同的使用顺序对最终结果的影响很微弱。
% 5. 在本文，我们选择“ocr-cap-den-tag”的顺序来进行实验。

\textbf{Impact of visual clue usage order on model performance.}
In the VIF module, we select one of four image-to-text models per iteration to extract visual clues from the given mention image, without repetition across iterations. In this experiment, we analyze the impact of the order in which visual clues are utilized on model performance. The experimental results are shown in Figure \ref{fig:sort_all}. We observe that the order of clue usage has minimal impact on model performance across the three datasets. The maximum performance difference between different orders is no more than 0.3\%. Based on this observation, we empirically choose the ``ocr-cap-den-tag'' order for our experiments in the study.

% , shown in Figure \ref{fig:sort_all}, demonstrate that the order of clue usage has minimal impact on the final outcomes. Based on these findings, we empirically choose the ``ocr-cap-den-tag'' order for our experiments in the study.

\textbf{Average response time of different LLM-based methods.}
To evaluate the efficiency of our framework, we introduce the average response time for selecting the Top-1 entity across each input mention in all samples as an additional metric, referred to as ``Avg-Time''. The comparison results with other LLM-based MEL methods (including UniMEL and GEMEL) are presented in Table \ref{tab:avg_time}. From the results, we observe the following: (1) Our proposed I2CR framework not only achieves a 3.4\% higher average accuracy compared to UniMEL, but also responds 3.27 seconds faster on average. 
This is because, for each sample, UniMEL requires at least two calls to the LLM and one to the MLLM. (2) Although our I2CR is slower than GEMEL in the Avg-Time metric, it achieves a notable 5.7\% higher average accuracy, significantly outperforming the GEMEL method.

\begin{table}[t]
\centering
\caption{Average response time (s) for returning the Top-1 result for all samples using different methods on each dataset. The values in the brackets at the lower right corner represent the accuracy improvement of I2CR compared to other LLM-based MEL methods (UniMEL and GEMEL).}
\scalebox{0.96}{
\begin{tabular}{>{\centering\arraybackslash}p{1.2cm} >{\centering\arraybackslash}p{1.4cm} >{\centering\arraybackslash}p{1.4cm} >{\centering\arraybackslash}p{1.4cm} >{\centering\arraybackslash}p{1.1cm}}
\toprule 
Model   & WikiMEL & WikiDiverse & RichMEL & Avg. \\ 
\midrule
I2CR                            & 8.51             & 6.63                 & 7.84             & 7.66          \\
\midrule
UniMEL                          & \multicolumn{1}{r}{10.97 \footnotesize{(3.5\%↑)}}    & \multicolumn{1}{r}{11.20 \footnotesize{(5.1\%↑)}}        & \multicolumn{1}{r}{10.61 \footnotesize{(1.6\%↑)}}    & \multicolumn{1}{r}{10.93 \footnotesize{(3.4\%↑)}} \\
GEMEL                           & \multicolumn{1}{r}{4.15 \footnotesize{(9.6\%↑)}}     & \multicolumn{1}{r}{4.53 \footnotesize{(5.3\%↑)}}         & -                & \multicolumn{1}{r}{4.34 \footnotesize{(5.7\%↑)}}  \\ \bottomrule
\end{tabular}
}
\label{tab:avg_time}
\end{table}

% \begin{table}[t]
% \centering
% \caption{The average response time of the samples}
% \scalebox{0.95}{
% \begin{tabular}{>{\centering\arraybackslash}p{1cm} >{\raggedleft\arraybackslash}p{0.6cm} >{\raggedright\arraybackslash}p{0.6cm} >{\raggedleft\arraybackslash}p{0.6cm} >{\raggedright\arraybackslash}p{0.6cm} >{\raggedleft\arraybackslash}p{0.6cm} >{\raggedright\arraybackslash}p{0.6cm} >{\raggedleft\arraybackslash}p{0.6cm} >{\raggedright\arraybackslash}p{0.6cm}}
% \toprule
% \multirow{2}{*}{Model}  & \multicolumn{8}{c}{The average response time of Top-1 (s)}                                                             \\ \cmidrule{2-9} 
%        & \multicolumn{2}{c}{WikiMEL} & \multicolumn{2}{c}{WikiDiverse} & \multicolumn{2}{c}{RichMEL} & \multicolumn{2}{c}{Avg.} \\ \midrule
% I2CR   & 8.51        &               & 6.63          &                 & 7.84        &               & 7.66      &              \\
% UniMEL & 10.97       & \footnotesize{(3.5\%↑)}      & 11.20         & \footnotesize{(5.1\%↑)}        & 10.61       & \footnotesize{(1.6\%↑)}      & 10.93     & \footnotesize{(3.4\%↑)}     \\
% GEMEL  & 4.15        & \footnotesize{(9.6\%↑)}      & 4.53          & \footnotesize{(5.3\%↑)}        & -           & -              & 4.34      & \footnotesize{(5.7\%↑)}     \\ \bottomrule
% \end{tabular}}
% \label{tab:avg_time}
% \end{table}

% 9. 为了更直观的理解每轮迭代视觉线索的重要性，我们提供了到第i（i=1，...，5）轮做对，前面i-1轮都做错的案例。
% 10. 当i=1时，第i-1轮的结果是第一个迭代中第一步生成的结果。
% 11. 从结果中我们发现每一轮迭代都能为mention找到合适的entity提供有效的帮助。
\textbf{Case study.}
To more intuitively understand the importance of visual clues in each iteration, we provide a case where the result in the $i$-th ($i = 1,...,5$) round is correct, while all prior rounds ($1$ to $i-$1) are incorrect. When $i$ = 1, the output of the ($i-$1)-th round refers to the result from the first step (the TES module) in the first iteration. The cases are shown in Figure \ref{fig:case}. Case (a) highlights the positive role of intra-modal and inter-modal collaborative verification. Cases (b) through (e) respectively validate the positive contributions of four different visual cues—OCR, Image Captioning, Dense Captioning, and Image Tagging—within the iterative process. In conclusion, each iteration round effectively contributes to identifying the correct entity for the given mention.

% 蓝色表示错误的输出，橙色表示正确的输出。

\section{Conclusion and Limitations}
In this study, we propose a novel LLM-based intra- and inter-modal collaborative reflection framework for the MEL task. The framework initially uses a fine-tuned LLM to select a candidate entity for the given mention based on mention text. If the entity selected from the text alone is determined to be incorrect through intra-modal consistency reflection and inter-modal alignment verification, we introduce a visual iterative feedback module. This module leverages visual clues generated by multiple image-to-text models for the mention image, assisting LLMs in refining its selection across several iterations to improve matching accuracy. Experimental results on WikiMEL, WikiDiverse, and RichMEL demonstrate that our proposed framework achieves SoTA performance, with additional detailed analyses validating the effectiveness of each component.

Although the proposed I2CR framework significantly improves model performance for the MEL task, it has two main limitations: 1) I2CR excels in handling common or general questions, but its effectiveness may be limited for long-tail questions (e.g., particularly rare mentions or entities). 2) I2CR is specifically designed for the textual and visual modalities in MEL, without considering other forms of data, such as speech or video, which may also contribute valuable information. We believe that with further optimization, our framework will evolve into a more comprehensive solution in the future.

% 1. 这一框架先使用文本来为mention选择候选实体。
% 2. 当通过intra-modal consistency reflection 和 inter-modal alignment verification判断仅用文本选择的实体不正确时，我们设计了visual iterative feedback模块来利用多个image-to-text models为mention image生成的描述来在不同迭代轮次中辅助模型，以提高匹配的精准性。
% 3. Experimental results on WikiMEL, WikiDiverse, and RichMEL demonstrate that our framework achieves SoTA performance, with additional detailed analyses that validate the effectiveness of each component.

\clearpage

% \section*{Limitations}
% Although the proposed I2CR framework significantly improves model performance for the MEL task, it has two main limitations: 1) I2CR excels in handling common or general questions, but its effectiveness may be limited for long-tail questions (e.g., particularly rare mentions or entities). 2) I2CR is specifically designed for the textual and visual modalities in MEL, without considering other forms of data, such as speech or video, which may also contribute valuable information. We believe that with further optimization, our framework will evolve into a more comprehensive solution in the future.

\begin{acks}
This paper was supported by the National Natural Science Foundation of China (No. 62306112), Shanghai Sailing Program (No. 23YF1409400), and Shanghai Pilot Program for Basic Research (No. 22TQ1400100-20).
\end{acks}
% \vspace{-2em}

\bibliographystyle{ACM-Reference-Format}
\balance
\bibliography{sample-base}

\end{document}